\documentclass[final]{article}
\usepackage{neurips_2023}
\usepackage[utf8]{inputenc} % allow utf-8 input
\usepackage[T1]{fontenc}    % use 8-bit T1 fonts
\usepackage{multirow}
\usepackage{hyperref}       % hyperlinks
\usepackage{url}            % simple URL typesetting
\usepackage{booktabs}       % professional-quality tables
\usepackage{amsfonts}       % blackboard math symbols
\usepackage{nicefrac}       % compact symbols for 1/2, etc.
\usepackage{microtype}      % microtypography
\usepackage{tabularx,colortbl}
\usepackage{graphicx}
\usepackage[skip=2pt,font=scriptsize]{caption}
\usepackage{float}
\usepackage{wrapfig}
\usepackage{natbib}
\setcitestyle{numbers,sort&compress, super}
\usepackage{subfig}% http://ctan.org/pkg/subfig
\newcommand{\sdbase}{SD-1.5-baseline}
\newcommand{\sdxlbase}{SDXL-baseline}
\newcommand{\sdxleg}{SDXL-DFT \textbf{with} ethnicity \& gender }
\newcommand{\sdxlnoeg}{SDXL-DFT \textbf{without} ethnicity \& gender }
\newcommand{\sdeg}{SD-1.5-DFT \textbf{with} ethnicity \& gender }
\newcommand{\sdnoeg}{SD-1.5-DFT \textbf{without} ethnicity \& gender }
\definecolor{Burgundy}{RGB}{144,0,32}
\usepackage[dvipsnames]{xcolor}
\title{Mitigating stereotypical biases in text to image generative systems}

\author{
Piero Esposito \quad Parmida Atighehchian \quad Anastasis Germanidis \quad Deepti Ghadiyaram \\
Runway \\
\texttt{\{pi, parmida, anastasis, deepti\}@runwayml.com}
}
\begin{document}
\maketitle
\begin{abstract}
State-of-the-art generative text-to-image models are known to exhibit social biases and over-represent certain groups like people of perceived lighter skin tones and men in their outcomes. In this work, we propose a method to mitigate such biases and ensure that the outcomes are \textit{fair} across different groups of people. We do this by finetuning text-to-image models on synthetic data that varies in perceived skin tones and genders constructed from diverse text prompts. These text prompts are constructed from multiplicative combinations of ethnicities, genders, professions, age groups, and so on, resulting in diverse synthetic data. Our \textit{diversity finetuned} (DFT) model improves the group fairness metric by $150\%$ for perceived skin tone and $97.7\%$ for perceived gender. Compared to baselines, DFT models generate more people with perceived darker skin tone and more women. To foster open research, we will release all text prompts and code to generate training images.

\end{abstract}
\section{Introduction}
\begin{wrapfigure}{r}{0.5\textwidth}
    \centering
    \vspace{-0.3in}
    \includegraphics[width=0.5\columnwidth]{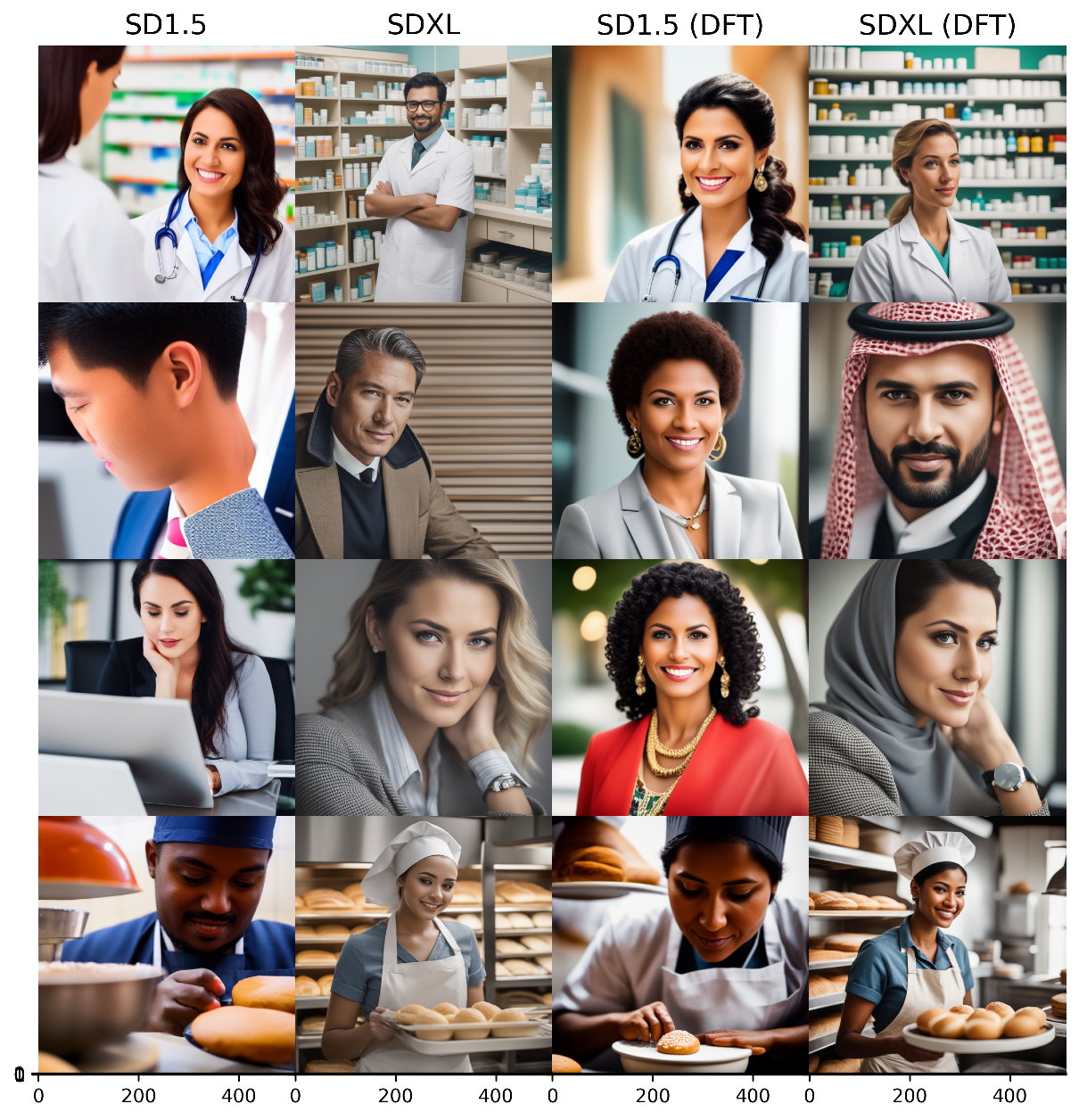}
    \vspace{-0.2in}
    \caption{We \scriptsize{\textbf{mitigate stereotypical biases} by finetuning Stable Diffusion-1.5~\cite{stable_diff} and Stable Diffusion-XL~\cite{sdxl} on synthetic data that varies across perceived skin tones, genders, professions, and age groups. For the same prompt and seed, notice that our diversity finetuned (DFT) models generate more inclusive results.}}
    \label{fig:fig1}
    \vspace{-0.2in}
\end{wrapfigure}

Generating high-quality photo-realistic images and videos from text is an active area of research~\cite{stable_diff, sdxl, imagen, dalle2, midjourney}, primarily fueled by the freely available (alt-text, image) pairs from HTML pages~\cite{laion_5b}. Despite the obvious benefits of generating complex visual content, training on billions of public web images brings forth inadvertent yet crucial challenges such as perpetuation of harmful social stereotypes and racial biases. Specifically, internet images contain amplified harmful stereotypes such as sexualized portrayal, especially of women~\cite{graff2013low} and under-representation of certain ethnicities and gender groups~\cite{steed2021image}. These could consequently seep into the trained models (Fig.~\ref{fig:fig1}). Given the widespread public availability of presumably biased text-to-image (TTI) models~\cite{sdxl, stable_diff, midjourney}, it is critical to measure and mitigate as many underlying harmful biases and stereotypes as possible. 

One way to alleviate these underlying social biases is to provide explicit qualifiers in the text prompts used for image generation. For example, if our goal is to correct for gender biases in certain professions, we could prompt a given TTI model to generate
`photo of a \textit{female} doctor' instead of just `photo of a doctor.' Yet, it has been shown in~\cite{stable_bias} that most TTI models~\cite{dalle2, stable_diff} still default to stereotypical representations despite providing explicit qualifiers. We tackle this issue in this work.

Our key idea is to supply a TTI model with more evidence of diverse representations of a given concept (eg: ``doctor'') during training so that the model implicitly adapts its latent space to capture a broader interpretation of that concept. We do this by focusing exclusively on finetuning on diverse training data, which is complementary to model-architectural adaptations~\cite{fair_diffusion}. Specifically, we construct diverse text prompts with multiplicative combinations of ethnicity, gender, age group, and professions (Table~\ref{tbl:template}). Then, we use an off-the-shelf TTI model to generate high-quality photo-realistic content from these prompts. Next, we finetune a given TTI model on this diverse generated data. At test time, we prompt the model \underline{without} any ethnicity, age, or gender based qualifiers. 

Given the difficulty in concretely measuring if stereotypical biases have been mitigated, we pose the evaluation in a \textit{group fairness} framework~\cite{disparate_impact} and measure if the generated results favor a stereotypical sub-group (e.g., perceived male, lighter skin tone). Through quantitative and qualitative evaluations, we show that our diversity finetuned (DFT) models achieve significant improvements in \textbf{not} defaulting to stereotypical representations and generating diverse content.  Beyond improving the overall capability of a model in generating more inclusive content, we explore training on compositions of different perceived skin tones and find that an equal skin tone representation during training yields maximum output diversity (Sec.~\ref{sec:data_comp}). We conduct user studies to understand if fine-tuning on synthetic data introduces any undesirable visual artifacts and find that users consistently prefer finetuned models' outputs. We also show that in addition to correcting for social biases, finetuning on high quality synthetic data also helps improve the overall output quality (Sec.~\ref{sec:output_quality}).

To summarize, our key contributions are: a) we generate diverse synthetic images of people from different professions, perceived skin tones and genders (Sec.~\ref{sec:data}), b) we finetune TTI models on them and improve the overall group fairness of Stable Diffusion~\cite{stable_diff} by $\mathbf{150\%}$ and Stable Diffusion-XL~\cite{sdxl} by $\mathbf{100\%}$ along perceived skin tone and $\mathbf{97.7\%}$ and $\mathbf{32.6\%}$ along perceived gender respectively (Sec.~\ref{sec:key_results}), c) we study the right training data composition and its impact on model fairness (Sec.~\ref{sec:data_comp}) and generation quality (Sec.~\ref{sec:output_quality}), d) to promote transparency and enable other researchers to also build less biased models, we will release all text prompts and code used for training image generation.
\section{Related work}
\textbf{Bias evaluation of TTI models} is relatively under explored compared to
other vision tasks like image captioning~\cite{women_snowboard, men_shop, race_caption} and face recognition~\cite{gender_shades}. Some recent works~\cite{tti_bias, tti_zhang} measure the behavior of TTI models when intervened by adding phrases about gender or skin tone to the original prompt. The authors of~\cite{dalleval} provide metrics for gender and skin tone inference and evaluation which do not require any prompt engineering; yet, we found the skin tone prediction to not generalize to diverse contents and the proposed MAD metric to not be very informative. Unlike all prior works~\cite{dalleval, stable_bias, tti_bias}, we uniquely formulate the evaluation in a group fairness~\cite{disparate_impact} framework. \\
\noindent \textbf{Finetuning} TTI models has been demonstrably very helpful for personalization~\cite{dreambooth, svdiff, emu}. Similar in spirit to our work, the authors of~\cite{overcoming_bias} show that finetuning on carefully curated data helps correct for some of the biases inherited during training for classical recognition tasks.
\section{Our Approach}
Our goal is to tweak a model's embeddings so that its generations do not default to certain stereotypical sub-groups like men or people of lighter skin tones. To this end, we finetune TTI models on a variety of synthetic images capturing multiplicative combinations of diverse perceived skin tones, genders, age groups, and professions, which we describe next.
\subsection{Data}\label{sec:data}

We start with the following template \textbf{<shot\_type> <age\_group> <ethnicity> <gender> <profession> <clothing> <location>} and generate text prompts from multiplicative combinations of the values for these qualifiers described in~Table~\ref{tbl:template}. Specifically, we source $57$ unique ethnicities and $170$ professions from Stereoset~\cite{stereoset} and Fair Diffusion~\cite{fair_diffusion}. We then make the template-generated prompts more descriptive using GPT-3.5~\cite{gpt3} (Appendix~\ref{sec:gpt3})), which allows us to synthesize images that capture more diversity in the background content, scene, and camera positions. This way, we get about $89K$ diverse prompts and use SDXL~\cite{sdxl} to produce $89K$ images to use for diversity finetuning. We note that we use \textbf{<ethnicity>} as a rough proxy for skin tones to generate images of people of diverse perceived skin tones. Of the $89K$ prompts, we have a roughly equal split of the two binary genders, male and female, with only $596$ prompts having a ``non-binary'' gender qualifier. 
\subsection{Fine-tuning setup}\label{sec:finetuning}
We finetune Stable Diffusion XL (SDXL)~\cite{sdxl} and Stable Diffusion-1.5 (SD-1.5)~\cite{stable_diff} (hyper-parameters in Table.~\ref{tbl:hyperparam}) under two settings: (a) \textbf{with ethnicity and gender qualifiers} in the text prompts (b) \textbf{without any ethnicity and gender qualifiers} in the text prompts. Our motivation is to investigate if adding explicit ethnicity and gender qualifiers to the text prompts further help a given TTI model in more broadly generalizing a concept -- like ``doctor'' -- across different perceived skin tones and genders. We stress that the \underline{same} qualifiers have been used to create the finetuning images.
\subsection{Evaluation setup} \label{sec:eval_setup}
At test time, we prompt the model in a neutral manner \underline{without} any ethnicity and gender based qualifiers. Each model is prompted on the same set of $340$ prompts and use $10$ seeds per prompt. Our goal is to determine if the model continues to default to certain subgroups or generates diverse results after finetuning. To this end, we use the disparate impact $80\%$ rule~\cite{disparate_impact}, which is a popular group fairness metric to determine the fairness in model outcomes. Let $X$ denote sub-groups like perceived genders (e.g., male, female) or skin tones (e.g., light, medium, and dark). If the ratio of the marginal probabilities of the \textit{predicted} occurrences between subgroup $X = female$ and $X = male$ is less than $\tau = 0.8$, this indicates that the underlying model favors $X = male$ over $X = female$, and is indicative of a biased model. We generalize this formulation for any given subgroups $X = \{X_1, X_2\}$: 
\begin{equation} \label{eqn:di}
Disparate \;Impact \; (DI) = {P(X' = X_1)}/{P(X' = X_2)}
\end{equation}
Under this setup, an ideal outcome is to have $DI \geq 0.8$, when $X_2$ denotes stereotypically favored sub-groups like perceived male or perceived lighter skin tone, for the model to not be biased. \\
%perceived skin tones and gender subgroups.\\ 
%We do this by finetuning on more diverse synthetic data.\\
\noindent \textbf{Perceived gender and skin tone predictions}
To predict perceived gender (male, female) and skin tone (light, medium, dark) from the generated images, we use BLIP-2~\cite{blip2} (Sec.~\ref{sec:blip2_prompts}). We compute disparate impact ratios and ablations on training data compositions (Sec.~\ref{sec:data_comp}) using these predictions.

\subsection{Ethical considerations} \label{sec:ethics}
Though we use ethnicity as a proxy for perceived skin tones, we in no way imply or believe that people of a particular ethnicity should have a particular skin tone. We found empirically that this way of constructing prompts yields diverse skin tone data, and helps generate more inclusive content. Though we use binarized socially-perceived gender labels, we note that this is not inclusive of all gender identities. Finally, to simplify our analysis, we group predicted skin tones into three buckets -- light, medium, and dark -- unlike $6$ in Fitzpatrick scale~\cite{gender_shades} or $10$ in Monk scale~\cite{monk1, monk2}.
\section{Experiments}
We first study the effect of qualifiers in the text prompts and then analyse training data composition. Since both \sdbase \,and \sdxlbase \,have very different architectures and pre-training datasets, throughout this section, we focus only on the relative improvement of the disparate impact metric (Eqn. ~\ref{eqn:di}). Since we compute $DI$ on BLIP-2's predictions (Sec.~\ref{sec:eval_setup}), it is important to first understand it's accuracy. For this, we manually label perceived skin tones and genders on $750$ random images and compare them with BLIP-2's predictions. We find that BLIP-2 achieves a very high classification accuracy of $99.9\%$ on gender but lower accuracy of $63.4\%$ on skin tone.
\begin{table}[h!]
\begin{center}
%\multirow{2}{*}{\shortstack[c]{\textbf{Semantic} \\ \textbf{Loss}}}
\scriptsize
\setlength{\tabcolsep}{0.51\tabcolsep}% Shrink \tabcolsep by 30%
\begin{tabular}{p{4cm}p{2cm}p{0.7cm}p{0.5cm}||p{1.6cm}p{0.5cm}}
\toprule
Model & L/M/D distribution & \multicolumn{2}{p{2cm}}{\textbf{Skin tone DI ($X_2 = light$)}} & F/M distribution & \multicolumn{1}{p{1.6cm}}{\textbf{Gender DI ($X_2 = male$)}} \\
\midrule
& & medium & dark &&  female \\
\midrule
\sdbase & 0.81 / 0.02 / 0.18 & {0.02} & {0.22} & 0.31 / 0.69 & {0.45} \\
\sdnoeg & 0.61 / 0.21 / 0.18 & 0.34 & 0.29 & 0.61 / 0.39 & \textcolor{ForestGreen}{1.57} \\
\sdeg & 0.57 / 0.12 / 0.32 & 0.21 & \textcolor{ForestGreen}{0.55} & 0.47 / 0.53 & \textcolor{ForestGreen}{0.89} \\
\midrule
\sdxlbase & 0.91 / 0.01 / 0.08 & {0.008} & 0.08 & 0.32 / 0.68 & {0.46} \\
\sdxlnoeg & 0.74 / 0.14 / 0.12 & {0.19} & {0.16} & 0.38 / 0.62 & \textcolor{ForestGreen}{0.61} \\
\sdxleg & 0.85 / 0.02 / 0.14 & {0.02} & {0.16} & 0.31 / 0.69 & {0.45} \\
\bottomrule
\end{tabular}
\vspace{0.03in}
\caption{\scriptsize{\textbf{Effect of prompt qualifiers} on group fairness. Given \textit{lighter skin tone} and \textit{perceived male gender} are the sub-groups TTI models default to, we measure disparate impact (Eqn.~\ref{eqn:di}) relative to these categories (i.e., $X_2 = {light, male}$). L/M/D refers to distribution of the predicted skin tones into \textbf{l}ight, \textbf{m}edium and \textbf{d}ark categories} and F/M refers to distribution into predicted \textbf{f}emale and \textbf{m}ale categories.} 
\label{tbl:main_result}
\vspace{-0.4in}
\end{center} 
\end{table}

\subsection{Effect of prompt qualifiers during training} \label{sec:key_results}
Here, we study the effect of finetuning with and without ethnicity and gender qualifiers in the text prompts using all $89$K images. From Table~\ref{tbl:main_result}, we note that for \sdbase, there is a significant improvement in the gender DI metric from $0.45$ to $\mathbf{0.89}$ and $\mathbf{1.57}$ when trained with and without qualifiers in the text prompts respectively. The F/M distribution further validates this improvement and indicates that training with qualifiers results in a more balanced outcome. Similar observations hold true for perceived skin tones, where, for dark skin tones, DI metric improves from $0.22$ to $0.55$ for \sdeg. Looking at the data distributions in both cases, we conclude that \sdeg yields a more balanced output. On the other hand, \sdxlnoeg yields more balanced results along both perceived skin tone and gender. Investigating the cause for this difference in behavior with regards to prompt modifiers for SD1.5 and SDXL is a fruitful future research direction. Finally, we note that the extent of improvement is more pronounced in SD1.5~\cite{stable_diff} compared to SDXL~\cite{sdxl}.
%\begin{table}[h!]
\begin{wraptable}{lt}{0.6\textwidth}
\vspace{-0.2in}
\begin{center}
\scriptsize
\setlength{\tabcolsep}{0.1\tabcolsep}
\begin{tabular}{p{4.5cm}p{2.1cm}p{0.9cm}p{0.5cm}}
\toprule
Model & L/M/D distribution & \multicolumn{2}{p{1.6cm}}{\textbf{Skin tone DI ($X_2 = light$)}}\\
\midrule
& & medium & dark \\
\midrule
\sdbase & 0.81 / 0.02 / 0.18 & {0.02} & {0.22} \\
\midrule
SD1.5-DFT + $11.9$K dark-only & 0.13 / 0.06 / 0.81 & 0.46 & 6.24 \\
SD1.5-DFT + $11.9$K medium-only & 0.66 / 0.32 / 0.01 & 0.49 & {0.02} \\
SD1.5-DFT + $11.9$K light-only & 0.82 / 0.12 / 0.06 & 0.15 & {0.08} \\
SD1.5-DFT + $35.6$K of light, medium, dark & 0.39 / 0.26 / 0.34 & {0.66} & {0.85} \\
%$89$K random & 0.61 / 0.21 / 0.18 & 0.34 & 0.29 \\
\bottomrule
\end{tabular}
\caption{\scriptsize{\textbf{Effect of data composition}. Note that a balanced distribution of perceived skin tones yields better performance. L/M/D refers to distribution of the predicted skin tones into \textbf{l}ight, \textbf{m}edium and \textbf{d}ark categories.}}
\label{tbl:skintone_composition}
\vspace{-0.6in}
\end{center} 
\end{wraptable}
\subsection{Effect of type of training data} \label{sec:data_comp}
Given the over-representation of lighter skin tone in the generated images of baseline models~\cite{stable_bias}, we study the optimal distribution of the perceived skin tones in the training data to improve group fairness. To this end, we construct data sets composed of only dark, medium, and lighter skin tones, and their combination (as predicted by BLIP-2~\cite{blip2}) and train independent SD1.5 models (without qualifiers) on these subsets. From the data distribution and the DI metrics in Table~\ref{tbl:skintone_composition}, it is clear that having a uniform distribution of perceived skin tones helps generate more diverse outputs. Compared to baseline, the DI metric improved from $0.02$ to $\mathbf{0.66}$ (by $\mathbf{3200}\%$) for medium and from $0.22$ to $\mathbf{0.85}$ (by $\mathbf{286.6\%}$) for dark skin tone when trained on fewer but more balanced data. Though adding only perceived dark skin tone data significantly increases generations with darker skin tone, it reduces the generations with light and medium skin tones and thus is not an ideal outcome.
\subsection{Impact on photo-realism, face deformations, and prompt alignment} \label{sec:output_quality}
Though generating synthetic data offers the advantage of scale and diverse content, we study if fine-tuning on such data introduces any unintended artifacts such as reduced photo-realism, poor quality faces, or prompt misalignment. In pair-wise subjective comparisons of generations from \sdbase \, and \sdnoeg, we observe that \sdnoeg was preferred $64.1\%$ and $55.1\%$ times over the baseline in terms of face quality and photo-realism respectively. We also report CLIP-scores~\cite{clip1, clip2} in a zero-shot setting on COCO~\cite{coco} for these two models and find the scores to be very stable ($0.2573$ v.s $0.2543$), indicating similar prompt alignment performance before and after finetuning. 
\section{Conclusions and Future work}
We propose a simple solution to mitigate social biases pertaining to perceived skin tone and gender in generative text to image models. We plan to continue addressing other forms of biases like income inequality, sexualized portrayals of women, and so on. We also wish to explore if image data can help mitigate biases in video models~\cite{gen1, alignlatents, pix2video, text2video} as well.
%in one of the two ways - a) engineer text prompts such that the visual concept of a “poor person” is not associated with a particular skin tone, and b) explicitly discourage the model during training by making it steer away from certain data distributions~\cite{concept_erasing, fair_diffusion} 
%While our finetuning technique demonstrates impressive capabilities in alleviating some common forms of social and racial biases, it does have some limitations. Firstly, our fine tuned model always generates a person of darker skin tone for certain prompts (e.g., “poor”). 
%We plan to investigate the cause behind relatively smaller improvements on SDXL~\cite{sdxl} (Sec.~\ref{sec:key_results}) compared to SD-1.5~\cite{stable_diff}. 
{\small 
\bibliographystyle{plain/unsrtnat}

}
%\bibliographystyle{splncs04}
%\bibliography{ref}
\section{Appendix}
\begin{figure}
%{r}{0.7\textwidth}
    \centering
    \vspace{-0.3in}
    \includegraphics[width=\columnwidth]{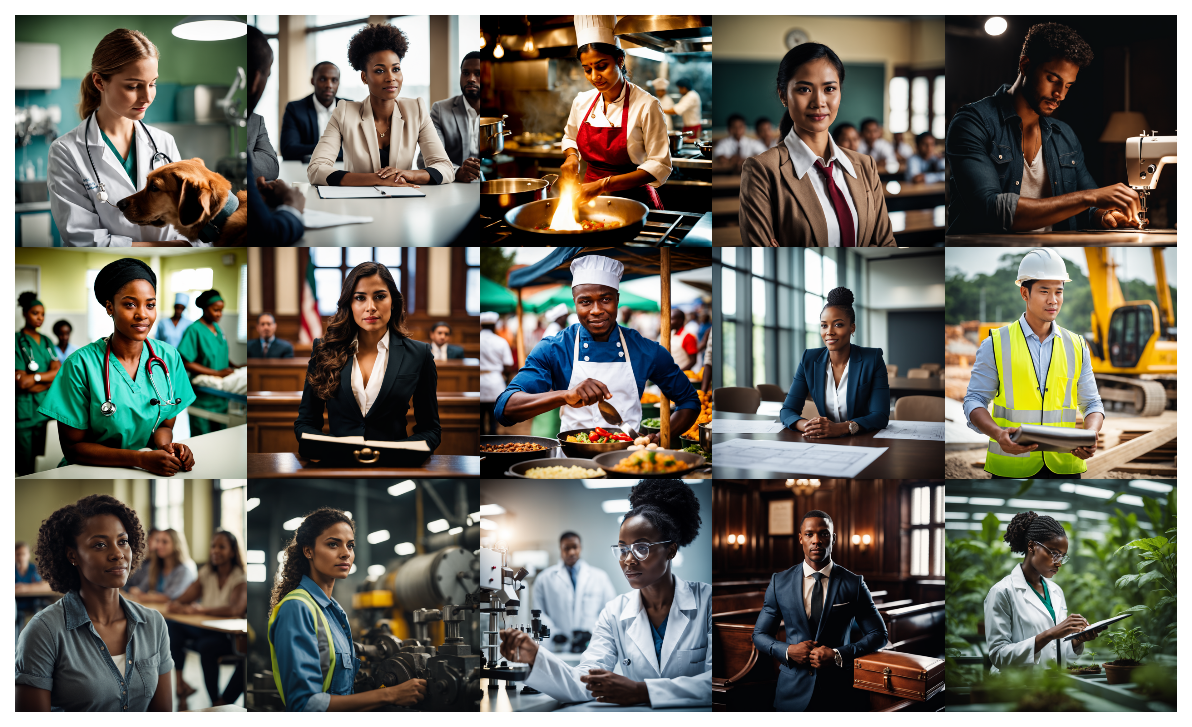}
  %  \vspace{-0.2in}
    \caption{\scriptsize{Example of training images we generated using paraphrased prompt templates described in Sec.~\ref{sec:data}}}
    \label{fig:fig1_backup}
%    \vspace{-0.2in}
\end{figure}
\begin{figure}
    \centering
    \includegraphics{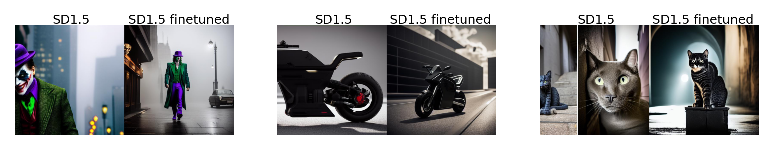}
    \vspace{-0.1in}
    \caption{\scriptsize{\textbf{Better visual quality outputs} with objects being centered are generated upon finetuning on high quality synthetic data.}}
    \label{fig:better_photo_realism}
    \vspace{-0.25in}
\end{figure}
\begin{figure}
    \centering
    \vspace{-0.3in}
    \includegraphics{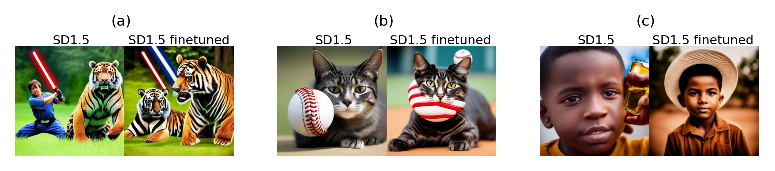}
    \vspace{-0.1in}
    \caption{\footnotesize{\textbf{Failure cases} Finetuning on synthetic data generated from SDXL~\cite{sdxl} leads to inheriting some issues of SDXL like concept bleeding ((a) and (b)) and reduced photo-realism or a waxy finish (c)}.}
    \label{fig:failure_cases}
    \vspace{-0.1in}
\end{figure}
%%%%%%%%%%%%%%%%%%%%%%%%%%%%%%%%%%%%%%%%%%%%%%%%%%%%%%%%%%
%\begin{wraptable}{l}{0.57\textwidth}
\begin{table}[t]
\setlength\extrarowheight{0.5pt}
\centering
\footnotesize
 \vspace{-1em}
\begin{tabular}{c | c }
\textbf{Qualifiers} & \textbf{Description} \\
\cmidrule{1-2}
shot type & close up, long shot, upper body shot \\
age & young, adult, old \\
gender & woman, man, non-binary \\
clothing & wearing work clothes, wearing ethnic clothes \\
location & at work, at home \\
professions & $170$ professions from Stereoset~\cite{stereoset} and Fair Diffusion~\cite{fair_diffusion}\\
ethnicity & $57$ unique ethnicities from Stereoset~\cite{stereoset} and Fair Diffusion~\cite{fair_diffusion}\\
\end{tabular}
\vspace{0.1in}
\caption{\footnotesize{We use the template \textbf{<shot\_type> <age\_group> <ethnicity>  <gender> <profession> <clothing> <location>}, where each qualifier takes a variety of values listed above. Using this technique, diverse text prompts of multiplicative combinations are generated.}} 
%\vspace{-0.5in}
\label{tbl:template}
\end{table}
\begin{table}[t]
\centering
\vspace{-0.1in}
\begin{tabular}{|c|c|c|}
\hline
Parameter & SD 1.5 & SDXL \\
\hline
\texttt{max\_train\_steps} & 3000 & 5000 \\
\texttt{mixed\_precision} & fp16 & fp16 \\
\texttt{learning\_rate} & \(3 \times 10^{-5}\) & \(1 \times 10^{-4}\) \\
\texttt{allow\_tf32} & true & true \\
\texttt{train\_batch\_size} & 16 & 6 \\
\texttt{gradient\_accumulation\_steps} & 2 & 4 \\
\texttt{lr\_scheduler} & polynomial & polynomial\\
\texttt{lr\_warmup\_steps} & 0 & 0\\
\texttt{resolution} & 512 & 768\\
\hline
\end{tabular} 
\caption{\footnotesize{\textbf{Hyper parameters} used while finetuning SD1.5 and SDXL.}} 
\label{tbl:hyperparam}
\vspace{-0.1in}
\end{table}
\begin{table}[h!]
%\begin{wraptable}{lt}{0.6\textwidth}
%\vspace{-0.2in}
\begin{center}
\footnotesize
\setlength{\tabcolsep}{0.1\tabcolsep}
\begin{tabular}{p{7.5cm}p{4.1cm}p{1.9cm}p{1.5cm}}
\toprule
Model & L/M/D distribution & \multicolumn{2}{p{3.6cm}}{\textbf{Skin tone DI ($X_2 = light$)}}\\
\midrule
& & medium & dark \\
\midrule
\sdbase & 0.81 / 0.02 / 0.18 & {0.02} & {0.22} \\
\sdnoeg  & 0.39 / 0.26 / 0.34 & {0.66} & {0.85} \\
\sdnoeg  & 0.46 / 0.09 / 0.45 & 0.21 & 0.98 \\ 
\midrule
\sdxlbase & 0.91 / 0.01 / 0.08 & {0.008} & {0.08} \\
\sdxleg  & 0.79 / 0.06 / 0.15 & 0.07 & 0.19 \\
\sdxlnoeg & 0.48 / 0.11 / 0.41 & 0.24 & 0.86 \\
\bottomrule
\end{tabular}
\vspace{0.06in}
\caption{\footnotesize{\textbf{Effect of data composition}. We train models with and without qualifiers on balanced skin tone data ($35.6$K of predicted light, medium, dark skin tone images). We observe that data composition plays a huge role in improvement of model fairness . L/M/D refers to distribution of the predicted skin tones into \textbf{l}ight, \textbf{m}edium and \textbf{d}ark categories.}}
\label{tbl:skintone_composition_appendix}
%\vspace{-0.6in}
\end{center} 
\end{table}
\subsection{GPT Instructions to diversify text prompts}\label{sec:gpt3}
\textit{You are a prompt generator. Given a sentence, you will paraphrase that sentence to create 2000 new prompts. The gender and the clothing of the person in the sentence should be different in each new prompt. The person should be always wearing cloth. The location and the job of the person and the camera position can be different in each prompt. The race of the person should be kept the same. The camera focus and camera shot should be changed in each prompt. Use creativity to add details in the prompts for the environment and the person.}
\subsection{Failure cases}
While our finetuning technique demonstrates impressive capabilities in alleviating some common forms of social and racial biases, it does have some inherent limitations. It is important to understand the limitations for future improvements. First, since we use SDXL~\cite{sdxl} to synthesize finetuning data, some of the limitations of SDXL such as concept bleeding are inherited by our model (Fig.~\ref{fig:failure_cases} (a) and (b)). In some cases, we noticed a reduction in photo realism (Fig.~\ref{fig:failure_cases} (c)). We plan to investigate and mitigate these issues in the future.

\subsection{Prompts used for inferring gender and skin tone} \label{sec:blip2_prompts}
\textbf{Prompt used to infer skin tone:} Answer the following question with `light` or `medium` or `black' (dark skin) or 'people not present' only. What is the skin tone of this person? \\
\textbf{Prompt used to infer gender:} Answer the following question with `male' or `female' or `people not present' only. Is this person on this file male or female?

\end{document}